\documentclass[11pt]{article}

\usepackage[final]{acl}

\usepackage{times}
\usepackage{latexsym}

\usepackage[T1]{fontenc}

\usepackage[utf8]{inputenc}

\usepackage{microtype}

\usepackage{inconsolata}

\usepackage{graphicx}
\usepackage{amssymb}
\usepackage{amsmath}
\usepackage{amsthm}
\usepackage{booktabs}
\usepackage{multirow}
\newtheorem{proposition}{Proposition}
\usepackage{hyperref}
\usepackage{color,soul}

\title{Text-to-Distribution Prediction with Quantile Tokens and Neighbor Context}

\author{
\textbf{Yilun Zhu}\textsuperscript{1}\thanks{Work done while at Amazon. Currently at Apple.} \quad
\textbf{Yuan Zhuang}\textsuperscript{1} \quad
\textbf{Nikhita Vedula}\textsuperscript{1} \quad
\textbf{Dushyanta Dhyani}\textsuperscript{1} \quad
\textbf{Shaoyuan Xu}\textsuperscript{1} \\
\textbf{Moyan Li}\textsuperscript{1} \quad
\textbf{Mohsen Bayati}\textsuperscript{2} \quad
\textbf{Bryan Wang}\textsuperscript{1} \quad
\textbf{Shervin Malmasi}\textsuperscript{1} \\
\\
\textsuperscript{1}Amazon.com, Inc. \qquad
\textsuperscript{2}Stanford University \\
\texttt{yz565@georgetown.edu} \quad
\texttt{bayati@stanford.edu} \\
\texttt{\{zyone, veduln, dhyanidd, shaoyux, moyanli, brywan, malmasi\}@amazon.com}
}

\begin{document}
\maketitle
\begin{abstract}
Many applications of LLM-based text regression require predicting a full conditional distribution rather than a single point value. We study \emph{distributional regression} under empirical-quantile supervision, where each input is paired with multiple observed quantile outcomes, and the target distribution is represented by a dense grid of quantiles. We address two key limitations of current approaches: the lack of local grounding for distribution estimates, and the reliance on shared representations that create an indirect bottleneck between inputs and quantile outputs. In this paper, we introduce \emph{Quantile Token Regression}, which, to our knowledge, is the first work to insert dedicated quantile tokens into the input sequence, enabling direct input-output pathways for each quantile through self-attention. We further augment these quantile tokens with retrieval, incorporating semantically similar \emph{neighbor} instances and their empirical distributions to ground predictions with local evidence from similar instances. We also provide the first theoretical analysis of loss functions for quantile regression, clarifying which distributional objectives each optimizes. Experiments on the Inside Airbnb and StackSample benchmark datasets with LLMs ranging from 1.7B to 14B parameters show that quantile tokens with neighbors consistently outperform baselines ($\sim$4 points lower MAPE and 2$\times$ narrower prediction intervals), with especially large gains on smaller and more challenging datasets where quantile tokens produce substantially sharper and more accurate distributions.\footnote{Our code is publicly available at \url{https://github.com/yilunzhu/text2distribution/}.}
\end{abstract}

\section{Introduction}
\label{sec:intro}

\begin{figure*}[t]
    \centering
    \includegraphics[width=\textwidth]{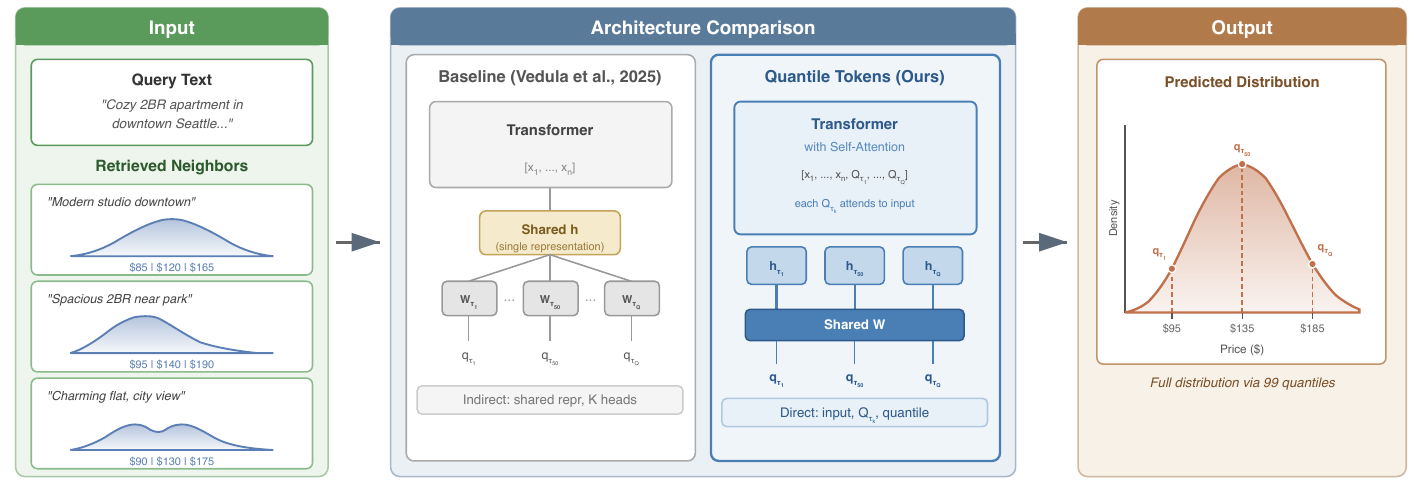}
    \caption{Overview of our approach. \emph{Left:} Input includes query text and retrieved neighbors with their full empirical distributions (as quantiles). \emph{Center:} The baseline \citep{vedula-etal-2025-quantile} uses only query text without neighbors and computes all quantiles from a shared hidden state via separate linear heads. Our Quantile Token approach augments the input with neighbors and inserts dedicated $\langle Q_\tau \rangle$ tokens that attend directly to the input, creating direct input-output pathways for each quantile. \emph{Right:} Output is a complete predicted distribution via $\tau$ quantiles. \emph{Bottom:} We evaluate on two diverse datasets (Airbnb, Stack Overflow) with Qwen3 1.7B--14B models.}
    \label{fig:overview}
\end{figure*}

Large Language Models (LLMs) have shown remarkable capabilities beyond text generation, extending to structured prediction tasks such as time series forecasting \citep{gruver2023large} and regression \citep{vacareanu2024wordsnumberslargelanguage, jacobs2024materials}. Recent work has shown that LLMs can approximate numerical mappings with strong accuracy when fine-tuned or prompted with in-context examples, making them attractive for text regression tasks where crucial information lies in unstructured text \citep{bitvai-cohn-2015-non, chen2024predicting}.

While most LLM-based regression work focuses on point estimation, many real-world use cases require predicting \emph{full probability distributions} rather than single values. Price prediction, demand forecasting, and risk assessment all benefit from understanding not just central tendencies but also dispersion and tail behavior \citep{arora2023probabilistic, kneib2023rage}. Quantile regression \citep{RePEc:ecm:emetrp:v:46:y:1978:i:1:p:33-50} provides a natural framework for distribution prediction by estimating conditional quantiles at different probability levels, offering robustness to outliers and the ability to capture heterogeneous effects across the distribution.

Recent work by \citet{vedula-etal-2025-quantile} takes an important first step towards LLM-based distributional prediction by attaching multiple linear regression heads to a shared final hidden state, each predicting a different quantile. However, this architecture has three key limitations. First, all quantile predictions derive from the same representation bottleneck, creating only an \emph{indirect} connection between input features and quantile-specific outputs. The model must compress everything relevant about the distribution into a single vector, from which separate heads attempt to extract different quantiles. Second, the method predicts distributions based only on the query text, which may contain limited information about the target distribution. This contrasts with how humans reason about distributions, which naturally relies on comparison with similar instances. For example, when estimating the price of a query product, one searches for similar products and builds an understanding by comparing features and observed price ranges. The model can benefit from a richer local context obtained by finding semantically similar items to inform the distribution estimate, but it does not have explicit access to such relevant reference points. Third, prior retrieval-augmented methods \citep{wang2025llpllmbasedproductpricing} rely on training with single point labels for each instance, which provides limited distributional supervision compared to training with full empirical distributions.

We address these limitations with a quantile token architecture augmented by neighbor context (Figure~\ref{fig:overview}). Our approach makes two key contributions:

\noindent \textbf{Quantile Token Regression.} We propose a novel architecture that inserts learnable quantile tokens ($\langle Q_{\tau_1}\rangle, \ldots, \langle Q_{\tau_Q}\rangle$) directly into the input sequence, which allows each quantile token to attend to different parts of the input and accumulate quantile-specific information. This creates a \emph{direct} input-output pathway for each quantile level, rather than relying on separate linear heads over a shared representation. The architecture enables more coherent quantile predictions since all quantile tokens are produced jointly within the same attention computation, and offers interpretability by revealing which input features each quantile attends to.

\noindent \textbf{Retrieval-Augmented Distribution Estimation.} We augment quantile regression with retrieved \textit{neighbor} instances, which are semantically similar examples from a candidate pool. Crucially, while prior retrieval-augmented approaches attach only a \emph{single point label} to each neighbor \citep{wang2025llpllmbasedproductpricing} and only predict a single price output, we equip each neighbor with its \emph{full empirical distribution} represented as quantiles. This provides the model with richer supervision. By grounding predictions in distributional evidence from neighbors, the model can better estimate not only central tendency but also dispersion and tail behavior of the target distribution.

To evaluate our approach, we construct two text-to-distribution datasets from Inside Airbnb~\citep{insideairbnb_get_the_data_2025} and StackSample~\citep{stackoverflow_stacksample_kaggle_2019,stackoverflow_content_license_2025}. Experiments with Qwen3 models~\citep{qwen3} spanning 1.7B--14B parameters show that (1) retrieval-augmented inputs consistently improve quantile regression across model scales (8\% relative reduction in avg MAPE on Airbnb, 63\% on StackSample), (2) quantile tokens outperform the shared-representation baseline (14\% relative reduction in avg MAPE on StackSample, 6$\times$ narrower intervals), and (3) the combination of both techniques yields the best performance. We also provide a mathematical analysis of different loss functions for quantile regression, clarifying the distributional objectives each optimizes. These two tasks capture different scales and uncertainty regimes for text-to-distribution prediction.

\section{Related Work}
\label{sec:related-work}

\paragraph{LLM-Based Regression.}
LLMs have been applied to regression through three main paradigms. First, in-context learning performs regression without fine-tuning by providing numeric examples in the prompt \citep{garg-etal-2022-function, vacareanu2024wordsnumberslargelanguage}. Second, LLM embeddings can be used as features for conventional regressors \citep{imperial2021bert, tang2025understanding}. Third, fine-tuning directly optimizes LLMs for numeric prediction, either by treating numbers as text tokens or by adding regression heads \citep{yang-etal-2020-enhancing, jacobs2024materials, song2024omnipred}. Recent work also explores regression-specific objectives, such as decision-theoretic fine-tuning (RAFT) \citep{lukasik2025raft} and coupling chain-of-thought with regression losses (TRACT) \citep{chiang-etal-2025-tract}. Our work follows the fine-tuning paradigm but targets \emph{distributional} rather than point prediction.

\paragraph{Quantile Regression and Distributional Prediction.}
Quantile regression \citep{RePEc:ecm:emetrp:v:46:y:1978:i:1:p:33-50} estimates conditional quantiles via pinball loss, enabling nonparametric characterization of predictive uncertainty. Distributional prediction is widely used in applications such as forecasting and risk-sensitive decision making \citep{arora2023probabilistic, gurlek2024boosted, gu2024predicting}, and recent work has begun integrating these ideas with LLMs \citep{gruver2023large, gillman2025fourier}. LLM-based quantile regression often attaches multiple quantile heads to a shared representation \citep{vedula-etal-2025-quantile, dorka2024qrm}, which can create a bottleneck between the input and quantile-specific outputs. In contrast, our quantile token architecture inserts dedicated tokens that participate in attention throughout the transformer, yielding more direct input--output pathways for each quantile.

\paragraph{Retrieval-Augmented Prediction.}
Retrieval augmentation has been effective for grounding LLM outputs in external evidence \citep{lewis2020retrieval, asai2023selfrag}. For regression, retrieved neighbors can provide context that improves calibration; for example, retrieval-augmented pricing leverages similar items to support numeric estimates \citep{wang2025llpllmbasedproductpricing}. Existing work largely targets point prediction. We extend retrieval augmentation to \emph{distributional} prediction, leveraging intuition that similar instances exhibit similar outcome distributions, which is informative for estimating dispersion and tail behavior.

\paragraph{Text Regression Applications.}
Text regression maps unstructured language to numeric targets in domains including finance, real estate, product pricing, and content scoring \citep{gu2024predicting, chen2024predicting, vedula-etal-2025-quantile, wang2025llpllmbasedproductpricing, chiang-etal-2025-tract}. A common challenge is that key signals are embedded in free-form text and are difficult to capture with manual features. We formulate text-to-distribution prediction as a general problem and evaluate across multiple domains (e.g., Airbnb listings and community Q\&A) to demonstrate breadth beyond pricing-centric settings.

\section{Method}

\subsection{Quantile Distribution Regression Task}
Let $X$ denote an input instance, where each input $X^{(i)}$ is a text sequence 
$X^{(i)}=(x^{(i)}_{1}, x^{(i)}_{2}, \ldots, x^{(i)}_{n_i})$, and let $Y\in\mathbb{R}$ be a continuous outcome. The objective is to learn a model that maps an input $X$ to a conditional distribution $F_{Y\mid X}(\cdot\mid X)$,
represented via its conditional quantiles. Specifically, for a set of quantile levels 
$\boldsymbol{\tau}=(\tau_1,\ldots,\tau_Q)\in(0,1)^Q$,
we predict the quantile vector 
$q_{\boldsymbol{\tau}}(X) = \big(q_{\tau_1}(X),\ldots,q_{\tau_Q}(X)\big)$,
where $q_{\tau}(X)$ is the $\tau$-th quantile of $F_{Y\mid X}(\cdot\mid X)$. This formulation follows the standard view of quantile regression as distribution learning through conditional quantiles.

\noindent \textbf{Empirical Quantile Supervision.}
Each input $X_i$ is paired with multiple observed outcomes
$\mathcal{Y}_i=\{y_{i1},\ldots,y_{iM_i}\}$, which we treat as realizations of $Y\mid X_i$.
We construct the empirical CDF
$\widehat{F}_i(t)=\frac{1}{M_i}\sum_{m=1}^{M_i}\mathbf{1}[y_{im}\le t]$
with empirical quantile function $\widehat{Q}_i(\tau)=\widehat{F}_i^{-1}(\tau)$.
Since each instance has a variable number of outcomes $M_i$, we interpolate $\widehat{Q}_i$ to a fixed grid of $Q=99$ quantile levels $\boldsymbol{\tau}=\{0.01,0.02,\ldots,0.99\}$ via linear interpolation (Appendix~\ref{app:quantile_interpolation}), producing target vectors $\widehat{\mathbf{q}}_i\in\mathbb{R}^{99}$.

\noindent \textbf{Learning Objective.}
Given a dataset $\mathcal{D}=\{(X_i,\mathcal{Y}_i)\}_{i=1}^N$, we train a model $f_\theta$ to predict $\widehat{\mathbf{q}}_i$ from $X_i$, i.e., $f_\theta(X_i)\approx \widehat{\mathbf{q}}_i$. This turns text-to-distribution prediction into structured regression over quantile levels.

\begin{figure*}[t!]
    \centering
    \includegraphics[width=\textwidth]{latex/figures/figure2_quantile_intuition_updated.pdf}
    \caption{Quantile tokens enable specialized representations and direct input-output pathways for each quantile level. For a Stack Overflow question about asyncio exception handling, different features signal different response times: the popular ``python'' tag suggests fast answers, while the niche ``asyncio'' topic and code complexity suggest slower responses. \emph{Center:} Each quantile token ($Q_{10}$, $Q_{50}$, $Q_{90}$) learns to attend to the features most predictive of its target quantile—$Q_{10}$ focuses on popularity signals while $Q_{90}$ focuses on complexity indicators. \emph{Right:} The resulting time-to-answer distribution captures both the possibility of a quick response (10\% within 15 minutes) and the long tail (90\% within 18 hours).} 
    \label{fig:quantile-intuition}
\end{figure*}

\subsection{Quantile Token Regression}
We introduce \emph{Quantile Token Regression}, a simple architectural change that makes each quantile prediction depend on a dedicated representation. Given an input sequence $X=(x_1,\ldots,x_n)$\footnote{Including the query and any retrieved neighbor context.}, we append $Q$ special quantile tokens $\langle Q_{\tau_1}\rangle,\ldots,\langle Q_{\tau_Q}\rangle$ to the end of the sequence. The resulting sequence is

{\small
\begin{equation}
\widetilde{X} = (x_1,\ldots,x_n,\langle Q_{\tau_1}\rangle,\ldots,\langle Q_{\tau_Q}\rangle).
\end{equation}
}
We feed $\widetilde{X}$ into a pretrained transformer $g_\theta$ and obtain final-layer hidden states $H=g_\theta(\widetilde{X})$.
Let $h_{\tau_k}\in\mathbb{R}^{d}$ denote the hidden state at the position of token $\langle Q_{\tau_k}\rangle$.
We then predict the $k$-th quantile using a shared linear regressor applied to the corresponding quantile-token representation.

{\small
\begin{equation}
\hat{q}_{\tau_k}(X) = w^\top h_{\tau_k} + b.
\end{equation}
}

Quantile token regression architecture has two advantages. 
First, it creates a direct input-output relation for each quantile level by allowing $\langle Q_{\tau_k}\rangle$ to collect information across all transformer layers, rather than relying on separate linear heads over a shared final representation. This improves the alignment between the conditioning evidence and quantile-specific predictions, which pays more attention to extreme quantiles. Figure~\ref{fig:quantile-intuition} illustrates this mechanism on a Stack Overflow question, where the range of response time to a question is predicted. The $\langle Q_{10}\rangle$ token learns to attend to popularity signals (e.g., ``python'' tag) that predict fast answers, while $\langle Q_{90}\rangle$ attends to complexity indicators that predict slower responses.
Second, quantile tokens result in more coherent quantile representations, since all quantile tokens are produced jointly within the same attention computation, and they enable interpretability by inspecting which parts of the input each $\langle Q_{\tau_k}\rangle$ attends to when forming its estimate.

\subsection{Retrieval-based Quantile Regression} \label{subsec:method_retrieval}
Semantically similar inputs tend to exhibit similar outcome distributions. For example, similar product descriptions yield similar price distributions, and similar questions receive similar response-time distributions. When conditioning only on the query text, the model must implicitly learn these distributional patterns from the training data, which can be challenging. We therefore augment quantile regression with \emph{neighbors}, semantically similar, label-bearing instances retrieved from a candidate pool, to explicitly provide the model with distributional evidence from similar instances.

Given an input text sequence $X$, we retrieve the top-$K$ semantically similar neighbors from a candidate pool using dense embedding similarity. Each retrieved neighbor is provided to the model along with its empirical distribution represented as a small set of quantiles. Implementation details, including the choice of embedding model, retrieval features, input formatting, and which quantiles are selected for neighbors, are described in Section~\ref{subsec:datasets}.

\subsection{Loss Functions and Theoretical Analysis}
\label{sec:loss_functions}

A critical design choice in distribution learning is the objective used to align the predicted quantile vector
$\hat{q}_{\boldsymbol{\tau}}(X)=\big(\hat{q}_{\tau_1}(X),\ldots,\hat{q}_{\tau_Q}(X)\big)$
with the supervision derived from the outcome set $\mathcal{Y}_i=\{y_{i1},\ldots,y_{iM_i}\}$.
Standard quantile regression applies the pinball loss to \emph{raw} outcomes $y$.
In contrast, our training targets $\widehat{Q}_i(\tau)$ are \emph{empirical quantile estimators} computed from a finite sample $\mathcal{Y}_i$ and then interpolated to a fixed grid.
We therefore analyze losses that are appropriate for \emph{quantile supervision}, and clarify what each objective targets in population.

\paragraph{$\ell_1$ and $\ell_2$ losses on empirical quantiles (Wasserstein matching).}
We treat the interpolated empirical quantiles $\widehat{Q}_i(\tau_k)$ as noisy measurements of the latent population quantiles $Q^*(X_i,\tau_k)$ and minimize an element-wise $\ell_p$ loss:

\vspace{-\baselineskip}
{\small
\begin{align}
\mathcal{L}_{\ell_1}(\theta) &= \frac{1}{N Q} \sum_{i=1}^N \sum_{k=1}^Q \left| \widehat{Q}_i(\tau_k) - \hat{q}_{\tau_k}(X_i) \right|, \\
\mathcal{L}_{\ell_2}(\theta) &= \frac{1}{N Q} \sum_{i=1}^N \sum_{k=1}^Q \left( \widehat{Q}_i(\tau_k) - \hat{q}_{\tau_k}(X_i) \right)^2.
\end{align}
}
This is called Wasserstein distance since in one dimension, the $p$-Wasserstein distance admits the quantile representation
$W_p^p(F,G)=\int_0^1 |Q_F(u)-Q_G(u)|^p\,du$ \citep{villani2009optimal}.
When $\boldsymbol{\tau}$ is a dense, approximately uniform grid, $\mathcal{L}_{\ell_1}$ and $\mathcal{L}_{\ell_2}$ can be interpreted as discrete approximations to
$W_1(\widehat{F}_i,\widehat{F}_\theta(\cdot\mid X_i))$ and $W_2^2(\widehat{F}_i,\widehat{F}_\theta(\cdot\mid X_i))$, respectively, where $\widehat{F}_\theta$ is the distribution implied by the predicted quantiles.

\paragraph{Mismatched pinball on empirical quantiles (Pinball-Q).}
An alternative is to apply the standard pinball loss to the empirical quantile targets:

\vspace{-\baselineskip}
{\small
\begin{equation}\label{eq:loss-Pinball-Q}
\mathcal{L}_{\text{Pinball-Q}}(\theta) = \frac{1}{N Q} \sum_{i=1}^N \sum_{k=1}^Q
\rho_{\tau_k}\!\left( \widehat{Q}_i(\tau_k) - \hat{q}_{\tau_k}(X_i) \right),
\end{equation}
}
where $\rho_{\tau}(u)=u\big(\tau-\mathbb{I}[u<0]\big)$.
While pinball is proper for learning $Q^*(X,\tau)$ from raw outcomes, here the ``outcome'' fed to pinball is itself a random quantile estimator.
As a result, \eqref{eq:loss-Pinball-Q} generally targets the $\tau$-th quantile of the \emph{estimator distribution} $\widehat{Q}_i(\tau)\mid X_i$, not the underlying parameter $Q^*(X_i,\tau)$ (Appendix~\ref{app:theory}).
This yields a systematic ``inflation/deflation'' effect away from $\tau=0.5$.

\paragraph{Scalarized pinball on a single statistic (Pinball-Med).}
Prior LLM regression work often associates each input with a single scalar label and trains all quantile heads against that scalar using pinball.
To mirror this setting under distribution supervision, we define a scalar pseudo-target
$y_i := \widehat{Q}_i(0.5)$ (the sample median)\footnote{If an application only provides a subset of quantiles, one can analogously set $y_i$ to the median of the available reported quantiles.}
and optimize

\vspace{-\baselineskip}
{\small
\begin{equation}\label{eq:loss-Pinball-Med}
\mathcal{L}_{\text{Pinball-Med}}(\theta) = \frac{1}{N Q} \sum_{i=1}^N \sum_{k=1}^Q
\rho_{\tau_k}\!\left( y_i - \hat{q}_{\tau_k}(X_i) \right).
\end{equation}
}
Appendix~\ref{app:theory} shows that \eqref{eq:loss-Pinball-Med} is generally \emph{inconsistent} for learning the full conditional distribution when $M_i>1$:
it learns the conditional distribution of the statistic $y_i$ (which concentrates around $Q^*(X_i,0.5)$ as $M_i$ grows).

\paragraph{Theoretical implications and empirical ordering.}
Under the latent-sample model in Appendix~\ref{app:theory}, for each fixed $(X,\tau)$ the empirical quantile obeys an asymptotic expansion
$\widehat{Q}_i(\tau) = Q^*(X_i,\tau) + \varepsilon_{i,\tau}$
where $\varepsilon_{i,\tau}$ is approximately centered and symmetric with variance scaling as
$\mathrm{Var}(\varepsilon_{i,\tau}\mid X_i)\propto \tau(1-\tau)/M_i$ (up to density factors).
Consequently, $\mathcal{L}_{\ell_1}$ and $\mathcal{L}_{\ell_2}$ yield Fisher-consistent estimations for $Q^*$ in the large-$M_i$ regime (with $\ell_1$ offering additional robustness under heavy tails or outliers).
In contrast, $\mathcal{L}_{\text{Pinball-Q}}$ introduces a bias of order $M_i^{-1/2}$ whose magnitude grows away from $\tau=0.5$, while $\mathcal{L}_{\text{Pinball-Med}}$ discards distributional shape and concentrates toward the median.
This analysis predicts the empirical ordering observed in our experiments:
$\mathcal{L}_{\ell_1}$ performs best overall, $\mathcal{L}_{\text{Pinball-Q}}$ is competitive but biased, and $\mathcal{L}_{\text{Pinball-Med}}$ underperforms due to systematic loss of tail information.

The same theory suggests variance-aware weighting by $(M_i,\tau_k)$ (and density factors) to downweight noisy tail quantiles when $M_i$ is small; we leave this extension to future work (Appendix~\ref{app:theory}).

\begin{table}[t]
\centering
\small
\setlength{\tabcolsep}{5pt}
\begin{tabular}{llrr}
\toprule
\textbf{Dataset} & \textbf{Split} & \textbf{\# Samples} & \textbf{Avg.\ \# Tokens} \\
\midrule
\multicolumn{4}{l}{\textbf{Inside Airbnb}} \\
 & train & 768{,}001 & 385 \\
 & val & 50{,}000 & 354 \\
 & test & 50{,}000 & 355 \\
 & test\_la (OOD) & 61{,}551 & 385 \\
\midrule
\multicolumn{4}{l}{\textbf{StackSample}} \\
 & train & 46{,}544 & 320 \\
 & val & 5{,}818 & 322 \\
 & test & 5{,}819 & 324 \\
\bottomrule
\end{tabular}
\caption{Dataset split statistics. We report number of samples and average number of tokens per sample.}
\label{tab:dataset}
\end{table}

\section{Experiments}

\begin{table*}[t!]
\centering
\small
\setlength{\tabcolsep}{4pt}
\renewcommand{\arraystretch}{1.05}
\resizebox{\textwidth}{!}{
\begin{tabular}{llr r r r r r r r}
\toprule
\textbf{Model} & \textbf{Method} & \textbf{$K$} &
\textbf{avg MAPE$\downarrow$} & \textbf{wMAPE$\downarrow$} & \textbf{sMAPE$\downarrow$} &
\textbf{CRPSS$\uparrow$} & \textbf{RCIW@90$\downarrow$} & \textbf{RCIW@95$\downarrow$} & \textbf{RCIW@99$\downarrow$} \\
\midrule
\multirow{3}{*}{Qwen3-1.7B} 
& Quantile Regression & 0 & 32.60 & 53.60 & 32.09 & 0.4408 & 12.72 & 16.29 & 19.06 \\
& Quantile Regression & 8 & 29.27 & 52.12 & 29.43 & 0.4588 & 16.73 & 22.80 & 29.84 \\
& Quantile Token & 8 & 27.18 & 50.18 & 27.39 & 0.4677 & 3.91 & 5.45 & 7.58 \\
\midrule
\multirow{3}{*}{Qwen3-4B} 
& Quantile Regression & 0 & 30.31 & 52.06 & 30.28 & 0.4536 & 9.58 & 12.30 & 14.66 \\
& Quantile Regression & 8 & 27.78 & 50.75 & 27.95 & 0.4700 & 11.77 & 15.08 & 17.94 \\
& Quantile Token & 8 & 26.89 & 49.99 & 27.14 & 0.4700 & 5.12 & 7.17 & 9.64 \\
\midrule
\multirow{3}{*}{Qwen3-8B} 
& Quantile Regression & 0 & 29.02 & 51.18 & 29.05 & 0.4616 & 7.61 & 9.69 & 11.34 \\
& Quantile Regression & 8 & 26.64 & 49.88 & 27.07 & \textbf{0.4800} & 7.38 & 9.48 & 11.27 \\
& Quantile Token & 8 & 26.56 & \textbf{49.63} & 26.75 & 0.4700 & \textbf{3.89} & \textbf{5.44} & \textbf{6.96} \\
\midrule
\multirow{3}{*}{Qwen3-14B} 
& Quantile Regression & 0 & 30.23 & 51.85 & 29.91 & 0.4575 & 10.89 & 14.25 & 16.87 \\
& Quantile Regression & 8 & 27.96 & 50.63 & 27.83 & 0.4754 & 17.46 & 22.98 & 27.51 \\
& Quantile Token & 8 & \textbf{26.40} & 49.67 & \textbf{26.67} & 0.4741 & 4.61 & 7.39 & 11.10 \\
\bottomrule
\end{tabular}
}
\caption{Results on the Airbnb test set with various model sizes. QR denotes quantile regression, QT denotes quantile token regression, and $K$ is the number of neighbors. Best value per metric column is in bold.}
\label{tab:airbnb_results}
\end{table*}
\begin{table*}[t]
\centering
\small
\setlength{\tabcolsep}{4pt}
\renewcommand{\arraystretch}{1.05}
\resizebox{\textwidth}{!}{
\begin{tabular}{llr r r r r r r r}
\toprule
\textbf{Model} & \textbf{Method} & \textbf{$K$} &
\textbf{avg MAPE$\downarrow$} & \textbf{wMAPE$\downarrow$} & \textbf{sMAPE$\downarrow$} &
\textbf{CRPSS$\uparrow$} & \textbf{RCIW@90$\downarrow$} & \textbf{RCIW@95$\downarrow$} & \textbf{RCIW@99$\downarrow$} \\
\midrule
Qwen3-4B & Quantile Regression & 0 & 266.65 & 75.00 & 67.41 & 0.0668 & 3900.57 & 6949.75 & 45480.31 \\
Qwen3-4B & Quantile Regression & 8 & 98.56  & 74.97 & 67.19 & 0.3001 & 545.50  & 1025.38 & 2110.01 \\
Qwen3-4B & Quantile Token & 8 & \textbf{84.30} & \textbf{73.02} & \textbf{64.86} & \textbf{0.3375} & \textbf{274.20} & \textbf{315.40} & \textbf{346.90} \\
\bottomrule
\end{tabular}
}
\caption{Results on the StackSample test split.}
\label{tab:stacksample_results}
\end{table*}

\subsection{Datasets} \label{subsec:datasets}
We experiment upon two publicly available text-to-distribution datasets from different domains: Inside Airbnb \citep{insideairbnb_get_the_data_2025} and StackSample \citep{stackoverflow_stacksample_kaggle_2019, stackoverflow_content_license_2025}. For both datasets, we construct ground-truth distributions from multiple observed outcomes per instance, keeping only instances with at least 4 observations (an empirical threshold chosen to balance label quality with dataset size). We use stratified sampling for train/val/test splits. For retrieval, we compute dense embeddings using \texttt{Qwen/Qwen3-Embedding-8B} over the full text of each instance. The retrieval candidate pool is restricted to the training split only, ensuring that no validation or test instances appear in retrieval. At inference time, each query retrieves top-$K$ neighbors (with $K{=}8$ in the main setting). Each retrieved neighbor contributes its title and nine representative empirical quantiles \footnote{1, 5, 10, 25, 50, 75, 90, 95, 99 percentiles.}, which are appended to the model input. For Airbnb, we build one index per city and perform same-city retrieval. For StackSample, the index is constructed from training questions only. For out-of-domain evaluation on Los Angeles (LA), LA listings are excluded from both training and retrieval candidate pools used in the main experiments. For LA evaluation, we construct a separate retrieval pool using only LA test instances, ensuring no cross-city information leakage. Dataset statistics are shown in Table~\ref{tab:dataset}; full construction details are in Appendix~\ref{app:datasets_details}.

\paragraph{Airbnb.}
The task is to predict the price distribution for an Airbnb listing given its textual description and metadata. We construct the dataset from Inside Airbnb~\citep{insideairbnb_get_the_data_2025}, collecting all available cities from 2024-09 to 2025-08 (119 cities) and converting prices to U.S.\ dollars. Each listing is represented by its title, description, amenities, location, and property type. We apply log transformation to prices to handle their wide range. We construct the ground-truth price distribution from observed monthly prices across time snapshots, yielding $\sim$840k samples from 55 cities after filtering. We hold out Los Angeles to form an out-of-domain (OOD) test set.

\paragraph{StackSample.}
The task is to predict the distribution of response times (time from question posting to receiving an answer) for a Stack Overflow question given its text. We use StackSample~\citep{stackoverflow_stacksample_kaggle_2019,stackoverflow_content_license_2025}, a Kaggle-hosted subset of Stack Overflow Q\&A. Each question is represented by its title, body, and tags. We construct the ground-truth response-time distribution from observed answer response times. We apply log transformation to handle the wide range of response times (from minutes to hours).

\subsection{Experimental Setup}
We fine-tune Qwen3 models (1.7B--14B parameters)~\citep{qwen3} with LoRA~\citep{hu2022lora}, predicting $Q{=}99$ uniformly spaced quantiles. We compare QR$_{K{=}0}$ (baseline quantile regression~\citep{vedula-etal-2025-quantile}), QR$_{K{=}8}$ (QR with $K{=}8$ retrieved neighbors), and QT$_{K{=}8}$ (quantile tokens with $K{=}8$ neighbors). We evaluate using average MAPE (Mean Absolute Percentage Error), wMAPE, sMAPE for point accuracy, and CRPSS (Continuous Ranked Probability Skill Score) and RCIW (Relative Coverage Interval Width) for distributional quality. Full details on dataset construction, experimental setup, and hyperparameters are in Appendix~\ref{app:experimental_setup} and~\ref{app:hyperparameters}. We focus on LLM-based quantile regression baselines as the strongest directly comparable approach under the same supervision and decoding setting. 
For both datasets, training is performed in log space; at inference, predictions are exponentiated back to the original scale before computing all evaluation metrics.

\subsection{Evaluation Metrics}
\label{subsec:metrics}

We evaluate quantile predictions on a test set $\{(x_i,y_i)\}_{i=1}^n$, where the model outputs $\hat{q}_{\tau}(x_i)$ for $\tau\in\{\tau_k\}_{k=1}^Q$. Since each sample has a ground-truth distribution, we first compute the Mean Absolute Percentage Error ($\mathrm{MAPE}$) at each quantile in a coarse set of target quantiles $\mathcal{T}=\{0.1,0.2,\ldots,0.9\}$. Then we compute the average over them, which we refer to as $\mathrm{average MAPE}$:

{\small
\begin{align}
\mathrm{MAPE}@\tau &= \frac{100}{n}\sum_{i=1}^n \left|\frac{\hat{q}_{\tau}(x_i)-q_{\tau}(x_i)}{q_{\tau}(x_i)}\right|, \\
\mathrm{avg MAPE} &= \frac{1}{|\mathcal{T}|}\sum_{\tau\in\mathcal{T}} \mathrm{MAPE}@\tau,
\end{align}
}

We additionally report the Weighted Mean Absolute Percentage Error ($\mathrm{wMAPE}$) and the Symmetric Mean Absolute Percentage Error ($\mathrm{sMAPE}$) using the median prediction $\hat{y}_i=\hat{q}_{0.5}(x_i)$. We use $y_i=q_{0.5}(x_i)$ as the benchmark value, which is the median of the ground-truth distribution.

{\small
\begin{align}
\mathrm{wMAPE} &= 100\cdot \frac{\sum_{i=1}^n |\hat{y}_i-y_i|}{\sum_{i=1}^n |y_i|}, \\
\mathrm{sMAPE} &= \frac{200}{n}\sum_{i=1}^n \frac{|\hat{y}_i-y_i|}{|\hat{y}_i|+|y_i|}.
\end{align}
}

For the evaluation of distributional quality, we report the Continuous Ranked Probability Skill Score (CRPSS) and Relative Coverage Interval Width (RCIW). CRPSS is

{\small
\begin{align}
\mathrm{CRPSS} = 1 - \frac{\mathrm{CRPS}_{\text{model}}}{\mathrm{CRPS}_{\text{ref}}},
\end{align}
}

\noindent where $\mathrm{CRPS}_{\text{ref}}$ is the CRPS of the empirical marginal distribution of training targets, following~\citet{vedula-etal-2025-quantile}. We report $\mathrm{RCIW}@c$ for $c\in\{90,95,99\}$:

{\small
\begin{align}
\mathrm{RCIW}@c = \frac{100}{n}\sum_{i=1}^n \frac{\hat{q}_{\tau_u(c)}(x_i)-\hat{q}_{\tau_\ell(c)}(x_i)}{\left|q_{0.5}(x_i)\right|},
\end{align}
}

\noindent where $(\tau_\ell,\tau_u)$ are chosen to give the closest available central coverage on our quantile grid (for $Q{=}99$, we use $(0.05,0.95)$, $(0.02,0.98)$, and $(0.01,0.99)$ for nominal $c{=}90,95,99$, respectively).

\begin{table*}[t]
\centering
\small
\setlength{\tabcolsep}{4pt}
\renewcommand{\arraystretch}{1.05}
\resizebox{\textwidth}{!}{
\begin{tabular}{llr r r r r r r r}
\toprule
\textbf{Model} & \textbf{Method} & \textbf{$K$} &
\textbf{avg MAPE$\downarrow$} & \textbf{wMAPE$\downarrow$} & \textbf{sMAPE$\downarrow$} &
\textbf{CRPSS$\uparrow$} & \textbf{RCIW@90$\downarrow$} & \textbf{RCIW@95$\downarrow$} & \textbf{RCIW@99$\downarrow$} \\
\midrule
\multirow{3}{*}{Qwen3-1.7B} 
& QR & 0 & 32.01 & 45.93 & 37.18 & 0.4706 & 14.06 & 17.98 & 21.05 \\
& QR & 8 & 24.95 & 37.68 & 25.64 & 0.5750 & 15.01 & 21.57 & 29.66 \\
& QT & 8 & 22.82 & 34.78 & 24.16 & 0.5901 & 3.72 & 5.18 & 7.35 \\
\midrule
\multirow{3}{*}{Qwen3-4B} 
& QR & 0 & 32.01 & 45.64 & 37.36 & 0.4680 & 10.00 & 12.76 & 14.92 \\
& QR & 8 & 23.54 & 35.79 & 24.30 & 0.5904 & 9.60 & 12.37 & 14.89 \\
& QT & 8 & 22.70 & 34.59 & 24.13 & 0.5952 & 5.10 & 7.12 & 9.64 \\
\midrule
\multirow{3}{*}{Qwen3-8B} 
& QR & 0 & 30.92 & 43.31 & 36.12 & 0.4950 & 8.26 & 10.52 & 12.29 \\
& QR & 8 & 22.56 & 34.32 & 23.30 & 0.6036 & 6.65 & 8.48 & 9.91 \\
& QT & 8 & \textbf{22.35} & 33.55 & 23.51 & 0.6058 & \textbf{3.76} & \textbf{5.16} & \textbf{6.58} \\
\midrule
\multirow{3}{*}{Qwen3-14B} 
& QR & 0 & 29.11 & 40.53 & 32.21 & 0.5332 & 11.54 & 14.72 & 17.20 \\
& QR & 8 & 23.33 & 35.23 & 24.50 & 0.6016 & 13.23 & 16.88 & 19.78 \\
& QT & 8 & 22.36 & \textbf{33.31} & \textbf{22.95} & \textbf{0.6128} & 5.38 & 8.41 & 11.70 \\
\bottomrule
\end{tabular}
}
\caption{Results on the Airbnb OOD test set (Los Angeles).}
\label{tab:airbnb_ood_results}
\end{table*}

\subsection{Experimental Results}
Our Quantile Tokens Regression approach (QT) consistently outperforms the quantile regression (QR) baseline across both datasets and all model sizes (Tables~\ref{tab:airbnb_results} and~\ref{tab:stacksample_results}). The method ranking is stable: QT$_{K{=}8}$ outperforms QR$_{K{=}8}$, which in turn outperforms QR$_{K{=}0}$ on all reported metrics. Compared to retrieval-augmented QR, QT improves both accuracy and sharpness, consistently reducing average MAPE and producing markedly narrower prediction intervals, with RCIW reduced by multiple factors in both datasets.\footnote{We note that StackSample exhibits substantially heavier-tailed distributions than Airbnb, leading to much larger absolute metric values (e.g., RCIW). Therefore, metric magnitudes are not directly comparable across datasets.} The advantage of QT is especially pronounced on StackSample, the smaller and more challenging dataset with response times spanning from 1.01 mins to 12 hrs. The QR$_{K{=}0}$ baseline produces extremely poor distributions with very wide confidence intervals (RCIW@99 of $4.55 \times 10^4$), while QT$_{K{=}8}$ converges more reliably. Comparing QT$_{K{=}8}$ to the baseline QR$_{K{=}0}$, QT$_{K{=}8}$ reduces average MAPE from 266.65 to 84.30\footnote{The absolute MAPE on StackSample remains high due to the limited dataset size and the inherently high uncertainty in response-time prediction compared to price prediction.} (68\% reduction) and shrinks RCIW@99 from $4.55 \times 10^4$ to 346.90 (131$\times$ reduction). Even compared to retrieval-augmented QR$_{K{=}8}$, QT$_{K{=}8}$ achieves 14\% lower average MAPE (84.30 vs 98.56) and 6$\times$ narrower intervals.

Neighbor retrieval provides substantial gains across all configurations, with particularly large impact when training data are limited. On StackSample, retrieval yields dramatic improvements: for the QR baseline, average MAPE drops from 266.65 to 98.56 (63\% reduction) when moving from $K{=}0$ to $K{=}8$. On Airbnb, retrieval also consistently improves average MAPE for QR across all model sizes: for Qwen3-4B, average MAPE drops from 30.31 to 27.78 (8\% reduction), and similar gains hold at other scales. These substantial improvements empirically validate the hypothesis (Section~\ref{subsec:method_retrieval}) that semantically similar inputs exhibit similar outcome distributions, as retrieved neighbors' distributions provide informative context for predictions. This contrast aligns with dataset scale: Airbnb is much larger, with $\sim$840k listings across 55 cities, while StackSample contains $\sim$58k questions. Therefore, retrieval has greater impact when training data are more limited, offering explicit distributional evidence from similar instances that helps ground predictions.

Model scaling shows diminishing returns at larger sizes on Airbnb. Moving from 1.7B to 4B parameters reduces average MAPE by 7\% for QT$_{K{=}8}$ (from 27.18 to 26.89), while moving from 8B to 14B yields only 1\% improvement (from 26.56 to 26.40). This aligns with empirical scaling laws~\citep{kaplan2020scalinglawsneurallanguage} but shows that gains saturate in our settings. Notably, distributional metrics do not monotonically improve with size: for example, QR$_{K{=}8}$ shows wider intervals at 14B than at 8B (RCIW@99 increases from 11.27 to 27.51). Since we tune hyperparameters per model, this suggests that larger backbones can be more sensitive to optimization and regularization choices, and better point accuracy does not necessarily translate into sharper confidence intervals.

To summarize, these results suggest that retrieval augmentation and quantile tokens are especially critical for harder, higher-uncertainty text-to-distribution tasks rather than providing only incremental gains. We further evaluate generalization by holding out Los Angeles in the Airbnb dataset during training and OOD testing on its listings, with full results in Appendix~\ref{app:exp_ood}.

\subsection{Experimental Results on OOD Dataset}
\label{app:exp_ood}

As shown in Table~\ref{tab:airbnb_ood_results}, the LA test set can appear easier than the multi-city test split for two complementary reasons. First, Los Angeles is a large, high-density city in the dataset, so retrieval can find closer and more informative neighbors, leading to stronger grounding for distribution prediction. Second, U.S.\ cities and listings constitute a substantial portion of our training data, so LA is not far from the dominant training distribution in both language and pricing patterns. As a result, the holdout primarily reflects a city-level split rather than a severe domain shift, which explains why OOD performance can be comparable to or even better than the stratified test set.

\subsection{Ablation Studies}

\subsubsection{Loss Functions}
Table~\ref{tab:ablation_loss} validates the theoretical comparison in Section~\ref{sec:loss_functions} and Appendix~\ref{app:theory} by contrasting Wasserstein losses with pinball variants under empirical-quantile supervision.

\begin{table}[h]
\centering
\small
\setlength{\tabcolsep}{5pt}
\renewcommand{\arraystretch}{1.05}
\begin{tabular}{lrrr}
\toprule
\textbf{Loss} & \textbf{avg MAPE$\downarrow$} & \textbf{CRPSS$\uparrow$} & \textbf{RCIW@95$\downarrow$} \\
\midrule
Pinball-Med            & 32.80 & \textbf{0.5331} & 151.78 \\
Pinball-Q   & 32.66 & \textbf{0.5332} & 151.27 \\
$\ell_1$ Wasserstein      & \textbf{26.55} & 0.4682 & \textbf{3.55} \\
$\ell_2$ Wasserstein     & 26.64 & 0.4737 & 4.15 \\
\bottomrule
\end{tabular}
\caption{Ablation on loss functions on Airbnb dev set using Qwen3-4B with $K{=}8$ neighbors.}
\label{tab:ablation_loss}
\end{table}

The two Wasserstein objectives, which are Fisher-consistent for the target quantiles as the number of labels per instance increases, achieve the best practical accuracy and sharpness. In particular, $\ell_1$ Wasserstein yields the lowest average MAPE and the tightest confidence intervals, while $\ell_2$ Wasserstein is competitive but slightly worse on both average MAPE and RCIW@95. In contrast, the pinball-based objectives perform poorly for distribution learning in our setting: \textsc{Pinball-Q} applies pinball loss to the empirical quantile targets, and \textsc{Pinball-Med} uses only the empirical median as supervision. Both incur much larger average MAPE and extremely wide intervals, consistent with the predicted bias of \textsc{Pinball-Q} and the loss of distributional information under \textsc{Pinball-Med}. Although the pinball losses attain higher CRPSS than Wasserstein, this is expected: CRPS is mathematically the integral of pinball losses across quantile levels~\citep{gneiting2007strictly}, so pinball training directly optimizes a discrete approximation of the evaluation criterion. However, this comes at the cost of massively inflated prediction intervals (RCIW@95 above 150 versus 3--4 for Wasserstein), rendering the resulting forecasts impractical despite the CRPSS advantage. Overall, these results support using Wasserstein objectives for empirical-quantile supervision, with $\ell_1$ Wasserstein providing the best accuracy--sharpness tradeoff in our experiments.

\subsubsection{Number of Neighbors}
Table~\ref{tab:ablation_neighbors} studies number of retrieved items $K$ on the Airbnb dev set using Qwen3-4B with QT and $\ell_1$ Wasserstein loss, applying postprocessed monotonicity (described in Section~\ref{sec:monotonicity}).

\begin{table}[h]
\centering
\small
\setlength{\tabcolsep}{6pt}
\renewcommand{\arraystretch}{1.05}
\begin{tabular}{crrr}
\toprule
\textbf{\# Neighbors} & \textbf{avg MAPE$\downarrow$} & \textbf{CRPSS$\uparrow$} & \textbf{RCIW@95$\downarrow$} \\
\midrule
0  & 29.54 & 0.4509 & 4.90 \\
2  & 27.02 & 0.4625 & 4.45 \\
4  & 26.64 & 0.4676 & 4.34 \\
8  & 26.55 & 0.4682 & 3.55 \\
16 & \textbf{25.85} & \textbf{0.4735} & \textbf{3.47} \\
\bottomrule
\end{tabular}
\caption{Ablation on the number of retrieved neighbors on Airbnb dev set using Qwen3-4B.}
\label{tab:ablation_neighbors}
\end{table}

Increasing $K$ consistently improves average MAPE and CRPSS while tightening intervals. The improvement is most pronounced when neighbors are first introduced, especially from $K{=}0$ to $K{=}2$, and exhibits diminishing returns as $K$ increases. Using $K{=}16$ performs best overall, reducing average MAPE from 29.54 to 25.85 and improving CRPSS from 0.4509 to 0.4735, while lowering RCIW@95 from 4.90 to 3.47. However, larger $K$ increases the number of input tokens, which raises memory usage and training cost (e.g., $K{=}16$ requires approximately 2$\times$ memory per sample and 1.4$\times$ total training time compared to $K{=}8$).

\subsubsection{Monotonicity}
\label{sec:monotonicity}

Our proposed Quantile Token regression approach provides no guarantee that the predicted quantiles will satisfy the monotonicity constraint, which can cause issues like the 90th percentile prediction being lower than the 80th percentile.
Table~\ref{tab:ablation_monotonicity} therefore compares three approaches to ensure monotonicity on the Airbnb dev set using Qwen3-4B with QT, $\ell_1$ Wasserstein loss, and $K{=}8$ neighbors.

\begin{table}[h]
\centering
\small
\setlength{\tabcolsep}{6pt}
\renewcommand{\arraystretch}{1.05}
\begin{tabular}{lrrr}
\toprule
\textbf{Method} & \textbf{avg MAPE$\downarrow$} & \textbf{CRPSS$\uparrow$} & \textbf{RCIW@95$\downarrow$} \\
\midrule
\textsc{baseline}    & 26.55 & 0.4682 & \textbf{3.55} \\
\textsc{cumsum}      & 26.51 & \textbf{0.4749} & 7.16 \\
\textsc{postprocess} & \textbf{26.35} & 0.4701 & \textbf{3.55} \\
\bottomrule
\end{tabular}
\caption{Ablation on monotonicity method on Airbnb dev set using Qwen3-4B with $K{=}8$ neighbors.}
\label{tab:ablation_monotonicity}
\end{table}

\textsc{Baseline} applies no monotonicity constraint and uses the raw predicted quantiles as-is. \textsc{Cumsum} enforces monotonicity during both training and inference by predicting non-negative gaps between adjacent quantiles and cumulatively summing them to form ordered quantiles. \textsc{Postprocess} keeps training unchanged and enforces monotonicity only at inference time by sorting the predicted quantiles. Empirically, \textsc{Cumsum} slightly improves average MAPE and yields the best CRPSS, but substantially widens intervals. In contrast, \textsc{Postprocess} achieves the lowest average MAPE, while maintaining comparable CRPSS and intervals.

\section{Conclusion}
We studied text-to-distribution prediction under empirical-quantile supervision, where each input has multiple observed outcomes and the target distribution is represented by a dense quantile grid. We introduced a \emph{retrieval-augmented} approach that grounds distribution estimates with retrieved neighbor instances and their empirical distributions, and \emph{Quantile Token Regression}, which predicts each quantile from a dedicated token representation formed through self-attention. Across Inside Airbnb and StackSample, both methods improve accuracy and yield sharper predictive intervals, with especially large gains on the smaller, more challenging StackSample dataset. We also analyzed training objectives, showing that Wasserstein matching better fits quantile-target supervision than pinball variants and offers a stronger accuracy--sharpness tradeoff in practice. Overall, combining retrieval-based grounding with quantile-specific representations is a simple, effective approach for scalable text-to-distribution prediction, motivating future work on variance-aware weighting, calibration under sharpness constraints, and broader applications.

\clearpage

\section*{Limitations}
Our evaluation relies on empirical quantiles constructed from multiple observed outcomes per input and interpolated to a fixed quantile grid; since ground-truth values at every quantile level are typically unavailable, this interpolation can introduce approximation error, especially when each instance has only a few labels, and the resulting distributions may not fully reflect real-world conditional outcome distributions. Our evaluation includes two LLM families, Qwen3 and Phi-3, but broader validation across additional backbones would further strengthen the generality of the findings. We report paired statistical significance tests, but do not study additional uncertainty estimates such as bootstrap confidence intervals or repeated data splits, which would further strengthen statistical reliability.

\bibliography{anthology,custom}

\appendix
\onecolumn

\section*{\center Appendix}
\label{sec:appendix}

\section{Theoretical Analysis of Quantile Supervision}
\label{app:theory}

This appendix formalizes the behavior of different objectives when training labels are \emph{empirical quantiles} computed from finite samples.
We justify the empirical ordering observed in our experiments:
$\mathcal{L}_{\ell_1}$ performs best, $\mathcal{L}_{\text{Pinball-Q}}$ is slightly worse but better than $\mathcal{L}_{\text{Pinball-Med}}$, and pinball-style objectives are appropriate only in the extremely sparse supervision regime.

\subsection{Latent sample model and quantile-label noise}

Fix an input instance $X=x$.
Let $F^*(\cdot\mid x)$ denote the true conditional distribution of $Y\mid X=x$ with (left-continuous) quantile function
\[
Q^*(x,\tau) := \inf\{t\in\mathbb{R} : F^*(t\mid x)\ge \tau\}, \qquad \tau\in(0,1).
\]
For simplicity, here we assume all $M_i$ are equal to $M$.
In our data, each input $X_i$ is paired with a multiset of outcomes $\mathcal{Y}_i=\{y_{i1},\ldots,y_{iM}\}$, which we model as
\[
y_{i1},\ldots,y_{iM}\ \overset{\text{i.i.d.}}{\sim}\ F^*(\cdot\mid X_i),
\]
with a variable sample size $M$ across instances.
From $\mathcal{Y}_i$ we construct the empirical CDF
$\widehat{F}_i(t)=\frac{1}{M}\sum_{m=1}^{M}\mathbb{I}[y_{im}\le t]$,
and define the empirical quantile estimator (with the same interpolation rule as in the main text)
\[
\widehat{Q}_i(\tau) := \widehat{F}_i^{-1}(\tau).
\]
For theoretical analysis, we treat $\widehat{Q}_i(\tau)$ as a standard sample quantile estimator; the additional linear interpolation changes the estimator by at most $O(1/M)$ when $f(Q)>0$ that does not change the asymptotics in the large $M$ regime.

\paragraph{Regularity assumption.}
We assume that for the quantile levels of interest $\tau\in[\tau_{\min},1-\tau_{\min}]$:
(i) $F^*(\cdot\mid x)$ is continuously differentiable in a neighborhood of $Q^*(x,\tau)$;
(ii) the conditional density $f^*(t\mid x)=\partial_t F^*(t\mid x)$ exists and satisfies
$f^*(Q^*(x,\tau)\mid x)>0$.

Under these conditions, sample quantiles admit a Bahadur-type expansion and a central limit theorem (e.g., ``Bahadur representation'', \cite{bahadur1966note}):
\begin{equation}\label{eq:bahadur}
\widehat{Q}_i(\tau)
=
Q^*(X_i,\tau)
+
\frac{\tau-\widehat{F}_i(Q^*(X_i,\tau))}{f^*(Q^*(X_i,\tau)\mid X_i)}
+
r_{i,\tau},
\qquad
r_{i,\tau}=o_p(M^{-1/2}).
\end{equation}
As a consequence, as $M$ grows to infinity,
\begin{equation}\label{eq:quantile_clt}
\sqrt{M}\Big(\widehat{Q}_i(\tau)-Q^*(X_i,\tau)\Big)
\ \Rightarrow\
\mathcal{N}\!\left(
0,\ 
\frac{\tau(1-\tau)}{f^*(Q^*(X_i,\tau)\mid X_i)^2}
\right).
\end{equation}
Thus $\widehat{Q}_i(\tau)$ can be viewed as a noisy measurement of $Q^*(X_i,\tau)$ with heteroskedastic noise that is (asymptotically) centered and symmetric.

\subsection{Population minimizers for $\ell_1$ and $\ell_2$ losses}

Because our loss functions sum over $k$, the linearity of expectation allows us to evaluate the population minimizers pointwise in $\tau$. Therefore, despite the finite-sample correlation across the empirical quantile vector, our proofs are pointwise in $\tau$ and rely only on the marginal distribution in \eqref{eq:quantile_clt}.

Fix $x$ and $\tau$ and define the random label $Z:=\widehat{Q}_i(\tau)\mid (X_i=x)$.
Consider a scalar prediction $a\in\mathbb{R}$.

\begin{proposition}\label{prop:bayes_l1_l2}
  The conditional risk minimizers satisfy:
  \begin{align*}
  \arg\min_{a\in\mathbb{R}}\ \mathbb{E}\big[(Z-a)^2\mid X=x\big] &= \mathbb{E}[Z\mid X=x], \\
  \arg\min_{a\in\mathbb{R}}\ \mathbb{E}\big[|Z-a|\mid X=x\big] &\in \mathrm{Median}(Z\mid X=x).
  \end{align*}
\end{proposition}


\begin{proof}
Both losses are convex in $a$.
For squared loss, differentiate the conditional risk and set to zero.
For absolute loss, the subdifferential of $a\to\mathbb{E}|Z-a|$ is 
\[
\partial = \left[2\mathbb{P}(Z<a|x)-1,2\mathbb{P}(Z\le a|x)-1\right]
\]
so $0\in\partial$ if and only if $\mathbb{P}(Z<a|x)\le \frac{1}{2}\le \mathbb{P}(Z\le a|x)$. This yields the median characterization.
\end{proof}

\begin{proposition}[Asymptotic Fisher consistency of $\mathcal{L}_{\ell_1}$ and $\mathcal{L}_{\ell_2}$]\label{prop:consistency_l1_l2}
Under the latent sample model and regularity assumptions above, for each fixed $(x,\tau)\in\mathcal{X}\times[\tau_{\min},1-\tau_{\min}]$, the population targets of $\mathcal{L}_{\ell_1}$ and $\mathcal{L}_{\ell_2}$ converge to the true quantile:
\[
\lim_{M\to\infty}
\mathbb{E}\!\left[\widehat{Q}_i(\tau)\mid X_i=x\right] = Q^*(x,\tau),
\qquad
\lim_{M\to\infty}\mathrm{Median}\!\left(\widehat{Q}_i(\tau)\mid X_i=x\right) = Q^*(x,\tau)\,.
\]
\end{proposition}

\begin{proof}
Fix an instance $X_i = x$ and a quantile level $\tau \in [\tau_{\min}, 1-\tau_{\min}]$. 

To establish convergence in mean, we use the Bahadur representation from Equation~\eqref{eq:bahadur}:
\[
\widehat{Q}_i(\tau) = Q^*(x,\tau) + \frac{\tau-\widehat{F}_i(Q^*(x,\tau))}{f^*(Q^*(x,\tau)\mid x)} + r_{i,\tau}\,.
\]
Taking the conditional expectation of both sides, we note that $\mathbb{E}[\widehat{F}_i(t)] = F^*(t\mid x)$. By definition, $F^*(Q^*(x,\tau)\mid x) = \tau$, meaning the expectation of the middle term is exactly zero. Under the stated regularity conditions, the expected value of the remainder term $r_{i,\tau}$ vanishes as $M \to \infty$. This gives the mean consistency:
\[
\lim_{M\to\infty} \mathbb{E}\!\left[\widehat{Q}_i(\tau)\mid X_i=x\right] = Q^*(x,\tau)\,.
\]

To establish median consistency, we rely on Equation~\eqref{eq:quantile_clt}, which shows that $\sqrt{M}(\widehat{Q}_i(\tau) - Q^*(x,\tau))$ converges in distribution to a centered Gaussian. Because the limiting normal distribution is continuous and strictly symmetric about zero, the median of the sequence of estimators converges to the median of the limiting distribution, yielding:
\[
\lim_{M\to\infty}\mathrm{Median}\!\left(\widehat{Q}_i(\tau)\mid X_i=x\right) = Q^*(x,\tau)\,.
\]

Finally, we bridge these asymptotic limits to our learning objectives via Proposition~\ref{prop:bayes_l1_l2}. Recall that for a given target variable $Z$, the population risk minimizers for squared error ($\mathcal{L}_{\ell_2}$) and absolute error ($\mathcal{L}_{\ell_1}$) correspond to the conditional expectation $\mathbb{E}[Z]$ and the conditional median $\mathrm{Median}(Z)$, respectively. Setting our target variable to the empirical quantile estimator, $Z = \widehat{Q}_i(\tau) \mid X_i=x$, and substituting the asymptotic limits established above, we obtain the limiting optimal predictions:
\begin{align*}
\lim_{M\to\infty} \left( \arg\min_{a\in\mathbb{R}}\ \mathbb{E}\big[(\widehat{Q}_i(\tau)-a)^2\mid X_i=x\big] \right) &= \lim_{M\to\infty} \mathbb{E}[\widehat{Q}_i(\tau)\mid X_i=x] = Q^*(x,\tau)\,, \\
\lim_{M\to\infty} \left( \arg\min_{a\in\mathbb{R}}\ \mathbb{E}\big[|\widehat{Q}_i(\tau)-a|\mid X_i=x\big] \right) &= \lim_{M\to\infty} \mathrm{Median}(\widehat{Q}_i(\tau)\mid X_i=x) = Q^*(x,\tau)\,.
\end{align*}
Consequently, as the number of observed outcomes $M \to \infty$, the population targets for both $\mathcal{L}_{\ell_1}$ and $\mathcal{L}_{\ell_2}$ correctly converge to the true latent quantile $Q^*(x,\tau)$, completing the proof.
\end{proof}

\subsection{Bias of mismatched pinball on empirical quantiles (Pinball-Q)}

\begin{proposition}\label{prop:bayes_pinball}
Let $\rho_\tau(u)=u(\tau-\mathbb{I}[u<0])$.
For any scalar random variable $Z$,
\[
\arg\min_{a\in\mathbb{R}}\ \mathbb{E}\big[\rho_\tau(Z-a)\big]
\ \in\ \mathrm{Quantile}_\tau(Z),
\]
i.e., pinball loss targets the $\tau$-th quantile of the label distribution.
\end{proposition}

\begin{proof}
Similar to the above proof for $\ell_1$, the subdifferential of $a\mapsto \mathbb{E}[\rho_\tau(Z-a)]$ is 
\[
\partial = \left[\mathbb{P}(Z<a|x)-\tau,\mathbb{P}(Z\le a|x)-\tau\right]
\]
so the optimality condition is $\mathbb{P}(Z<a|x)\le \tau\le \mathbb{P}(Z\le a|x)$.
\end{proof}

\begin{proposition}[Bias of $\mathcal{L}_{\text{Pinball-Q}}$]\label{prop:bias_pinq}
Fix $(x,\tau)$ and write the empirical quantile label as
\[
\widehat{Q}_i(\tau)=Q^*(x,\tau)+\varepsilon_{i,\tau},
\qquad
\varepsilon_{i,\tau}:=\widehat{Q}_i(\tau)-Q^*(x,\tau).
\]
The population minimizer of $\mathcal{L}_{\text{Pinball-Q}}$ at level $\tau$ satisfies
\[
q^*_{\text{Pinball-Q}}(x,\tau)
=
Q^*(x,\tau)
+
\mathrm{Quantile}_\tau\!\big(\varepsilon_{i,\tau}\mid X_i=x\big).
\]
Under the normal approximation in \eqref{eq:quantile_clt}, the resulting inflation/deflation bias is approximately
\begin{equation}\label{eq:pinq_bias_normal}
q^*_{\text{Pinball-Q}}(x,\tau)
\approx
Q^*(x,\tau)
+
\Phi^{-1}(\tau)\,
\sqrt{\frac{\tau(1-\tau)}{M}}\,
\frac{1}{f^*(Q^*(x,\tau)\mid x)}.
\end{equation}
\end{proposition}

\begin{proof}
By Proposition~\ref{prop:bayes_pinball}, $\mathcal{L}_{\text{Pinball-Q}}$ targets the $\tau$-quantile of the random label $\widehat{Q}_i(\tau)\mid X_i=x$.
Quantiles are translation-equivariant, giving the first display.
Under \eqref{eq:quantile_clt}, $\varepsilon_{i,\tau}$ is approximately normal with standard deviation
$\sqrt{\tau(1-\tau)}/(\sqrt{M}\,f^*(Q^*\mid x))$, and the $\tau$-quantile of a centered normal is $\Phi^{-1}(\tau)$ times its standard deviation.
\end{proof}

Equation~\eqref{eq:pinq_bias_normal} shows that the leading $O(M^{-1/2})$ inflation term vanishes at $\tau=0.5$, but Pinball-Q systematically inflates upper quantiles ($\tau>0.5$) and deflates lower quantiles ($\tau<0.5$).
The magnitude of this mismatch is $O(M^{-1/2})$ and grows away from $\tau=0.5$ over typical interior grids and can be further amplified when $f^*$ is small (often in tails).

\subsection{Scalarized pinball is inconsistent for distribution learning (Pinball-Med)}

Define the scalar pseudo-target
\[
y_i := \widehat{Q}_i(0.5),
\]
and recall the scalarized objective
$\mathcal{L}_{\text{Pinball-Med}}(\theta)=\frac{1}{NQ}\sum_{i,k}\rho_{\tau_k}(y_i-\hat q_{\tau_k}(X_i))$.

\begin{proposition}[What $\mathcal{L}_{\text{Pinball-Med}}$ learns]\label{prop:pinmed_target}
For each $\tau\in(0,1)$ and fixed $x$, the population minimizer of $\mathcal{L}_{\text{Pinball-Med}}$ at level $\tau$ satisfies
\[
q^*_{\text{Pinball-Med}}(x,\tau) \in \mathrm{Quantile}_\tau\big(y_i \mid X_i=x\big).
\]
Consequently, $\mathcal{L}_{\text{Pinball-Med}}$ is proper for the conditional distribution of the statistic $y_i=\widehat{Q}_i(0.5)$, not for $Y\mid X$.
\end{proposition}

\begin{proof}
Fix $(x,\tau)$ and view $y_i$ as the random label.
Applying Proposition~\ref{prop:bayes_pinball} to $Z=y_i\mid X_i=x$ yields the claim.
\end{proof}

\begin{proposition}[Concentration in the large-$M$ regime]\label{prop:pinmed_collapse}
Assume $M\to\infty$ and the regularity conditions above.
Then $y_i=\widehat{Q}_i(0.5)\to Q^*(x,0.5)$ in probability, and hence
\[
q^*_{\text{Pinball-Med}}(x,\tau)\to Q^*(x,0.5)\qquad\text{for all }\tau\in(0,1).
\]
If $F^*(\cdot\mid x)$ is non-degenerate, this implies a non-vanishing distributional error; for example,
\[
W_1\!\left(F^*(\cdot\mid x),\ \delta_{Q^*(x,0.5)}\right)
=
\int_0^1 \big|Q^*(x,u)-Q^*(x,0.5)\big|\,du
\;>\;0.
\]
\end{proposition}

\begin{proof}[Proof sketch]
Consistency of the sample median follows from standard quantile consistency.
As $y_i$ concentrates, the conditional distribution of $y_i\mid X_i=x$ converges to a point mass at $Q^*(x,0.5)$, whose $\tau$-quantile equals the same point for every $\tau$.
The $W_1$ identity is standard in one dimension.
\end{proof}



\subsection{Towards a variance-aware weighting}
Equation~\eqref{eq:quantile_clt} implies heteroskedastic noise across $(M,\tau)$.
A variance-aware extension replaces the unweighted losses with weights proportional to the inverse asymptotic variance,
\[
w_{i,k}\ \propto\ \frac{M\, f^*(Q^*(X_i,\tau_k)\mid X_i)^2}{\tau_k(1-\tau_k)},
\]
yielding a quasi-likelihood weighted least squares objective.
In practice, the unknown density factor can be estimated from the predicted quantile slope via
$f^*(Q^*(x,\tau)\mid x)=1/\partial_\tau Q^*(x,\tau)$ (when the quantile function is differentiable and strictly increasing), suggesting a fully data-adaptive weighting scheme.
We leave a systematic study of these weights to future work.

\section{Quantile Interpolation}
\label{app:quantile_interpolation}

Each instance is associated with a variable-size set of observed outcomes $\mathcal{Y}_i=\{y_{i1},\ldots,y_{iM_i}\}$, where $M_i$ can differ across instances. We sort them to obtain order statistics $y_{i(1)}\le \cdots \le y_{i(M_i)}$ and treat these samples as defining an empirical quantile function. To obtain a fixed-dimensional training target, we interpolate the empirical quantile function to a dense grid of $Q=99$ quantile levels $\boldsymbol{\tau}=\{0.01,0.02,\ldots,0.99\}$, producing

{\small
\begin{equation}
\widehat{\mathbf{q}}_i=\big(\widehat{Q}_i(0.01),\ldots,\widehat{Q}_i(0.99)\big)\in\mathbb{R}^{99}.
\end{equation}
}

For each $\tau_k$, we compute a fractional rank $r_{ik} = 1 + (M_i-1)\tau_k$ and set $\widehat{Q}_i(\tau_k)$ by linear interpolation between $y_{i(\lfloor r_{ik}\rfloor)}$ and $y_{i(\lceil r_{ik}\rceil)}$.

\section{Dataset Construction Details}
\label{app:datasets_details}

\paragraph{Airbnb.}
We collect all available cities from Inside Airbnb data snapshots between 2024-09 and 2025-08, covering 119 cities. We drop listings with fewer than 4 price observations and remove cities with fewer than 10k samples after filtering, yielding $\sim$840k samples from 55 cities. For retrieval, we build one training index per city and restrict retrieved neighbors to the same city. For each retrieved neighbor, we append only its title to the model input.

\paragraph{StackSample.}
We filter out answers with response time exceeding 12 hours, convert each remaining response time to minutes, then apply log transformation to ensure a manageable range. We perform quantile interpolation over the log-transformed response times to create the ground-truth quantile distribution. For retrieval, we build an index from training questions only. For each retrieved neighbor question, we append only its title to the model input.

\section{Experimental Details}
\subsection{Experimental Setup}
\label{app:experimental_setup}

We run all experiments on a single AWS GPU cluster with NVIDIA H100 and H200 GPUs, set $Q=99$ with uniformly spaced quantile levels, and tune hyperparameters (e.g., number of epochs, batch sizes, learning rates, and maximum sequence length) across settings, model sizes, and datasets to report the best configuration. Unless otherwise specified, we use $K=8$ retrieved neighbors. We fine-tune the Qwen3 model family \citep{qwen3} with LoRA \citep{hu2022lora} using the HuggingFace Transformers library \citep{wolf2020huggingfacestransformersstateoftheartnatural}.

\subsection{Hyperparameters}
\label{app:hyperparameters}

This section provides the hyperparameter configurations used in our experiments. All models were trained using LoRA fine-tuning on the Qwen3 family. Table~\ref{tab:airbnb_hyperparams} shows hyperparameters for Inside Airbnb experiments, and Table~\ref{tab:stacksample_hyperparams} shows hyperparameters for StackSample experiments.

\begin{table*}[tbh]
\centering
\small
\setlength{\tabcolsep}{4pt}
\begin{tabular}{llrrrrrrr}
\toprule
\textbf{Model} & \textbf{Method} & \textbf{K} & \textbf{LR} & \textbf{Epochs} & \textbf{Train BS} & \textbf{Eval BS} & \textbf{LoRA Rank} & \textbf{Max Len} \\
\midrule
Qwen3-1.7B & QR & 0 & 3e-6 & 5 & 64 & 64 & 384 & 1024 \\
Qwen3-1.7B & QR & 8 & 3e-6 & 5 & 64 & 64 & 384 & 2048 \\
Qwen3-1.7B & QT & 8 & 3e-6 & 5 & 16 & 8 & 384 & 2048 \\
\midrule
Qwen3-4B & QR & 0 & 2e-6 & 5 & 32 & 32 & 384 & 1024 \\
Qwen3-4B & QR & 8 & 2e-6 & 5 & 32 & 32 & 384 & 2048 \\
Qwen3-4B & QT & 8 & 2e-6 & 5 & 16 & 16 & 384 & 2048 \\
\midrule 
Qwen3-8B & QR & 0 & 2e-6 & 5 & 16 & 16 & 384 & 1024 \\
Qwen3-8B & QR & 8 & 2e-6 & 5 & 16 & 16 & 384 & 2048 \\
Qwen3-8B & QT & 8 & 2e-6 & 5 & 16 & 16 & 384 & 2048 \\
\midrule
Qwen3-14B & QR & 0 & 1e-6 & 5 & 32 & 32 & 384 & 1024 \\
Qwen3-14B & QR & 8 & 1e-6 & 5 & 16 & 16 & 384 & 2048 \\
Qwen3-14B & QT & 8 & 1e-6 & 5 & 8 & 8 & 384 & 2048 \\
\bottomrule
\end{tabular}
\caption{Hyperparameters for Inside Airbnb experiments. K = number of neighbors, LR = learning rate, BS = batch size per device, Max Len = maximum sequence length. All models use weight decay = 0.01, LoRA dropout = 0.1, and Wasserstein W1 loss.}
\label{tab:airbnb_hyperparams}
\end{table*}

\begin{table*}[htb]
\centering
\small
\setlength{\tabcolsep}{4pt}
\begin{tabular}{llrrrrrrrrr}
\toprule
\textbf{Model} & \textbf{Method} & \textbf{K} & \textbf{LR} & \textbf{Epochs} & \textbf{Train BS} & \textbf{Eval BS} & \textbf{LoRA Rank} & \textbf{WD} & \textbf{Warmup} & \textbf{Max Len} \\
\midrule
Qwen3-4B & QR & 0 & 3e-7 & 8 & 32 & 64 & 64 & 0.01 & 0.0 & 1280 \\
Qwen3-4B & QR & 8 & 3e-7 & 8 & 32 & 32 & 64 & 0.01 & 0.0 & 2304 \\
Qwen3-4B & QT & 8 & 3e-7 & 5 & 16 & 16 & 192 & 0.01 & 0.1 & 2304 \\
\bottomrule
\end{tabular}
\caption{Hyperparameters for StackSample experiments. K = number of neighbors, LR = learning rate, BS = batch size per device, WD = weight decay, Warmup = warmup ratio, Max Len = maximum sequence length. QR models use LoRA dropout = 0.1, QT model uses LoRA dropout = 0.15. All use Wasserstein W1 loss.}
\label{tab:stacksample_hyperparams}
\end{table*}

\subsection{Retrieval Efficiency and Cost}
\label{app:retrieval_cost}

While our main results and ablations show that retrieval augmentation consistently improves predictive performance, incorporating retrieved context increases input length and computational cost. Table~\ref{tab:retrieval_cost} quantifies this tradeoff for Qwen3-4B on the Airbnb dataset.

\begin{table}[htb]
\centering
\small
\setlength{\tabcolsep}{5pt}
\begin{tabular}{rrrr}
\toprule
$K$ & Train Time (hrs) & Train Speed (samples/s) & Inference Speed (samples/s) \\
\midrule
0  & 23.71 & 44.99 & 208.52 \\
2  & 32.69 & 32.63 & 150.06 \\
4  & 42.68 & 25.00 & 115.54 \\
8  & 64.95 & 16.42 & 76.43  \\
16 & 90.32 & 11.81 & 54.48  \\
\bottomrule
\end{tabular}
\caption{Training and inference efficiency as the number of retrieved neighbors $K$ increases (Qwen3-4B on Airbnb).}
\label{tab:retrieval_cost}
\end{table}

As the number of retrieved neighbors $K$ increases, both training and inference throughput decrease substantially. Increasing $K$ from 0 to 8 raises training time from 23.71 to 64.95 hours (2.7$\times$) and reduces inference throughput from 208.52 to 76.43 samples/sec. Combined with the accuracy results in Table~\ref{tab:ablation_neighbors}, these measurements suggest $K{=}8$ as a practical operating point that balances predictive gains against computational cost.

\subsection{Statistical Significance}
\begin{table*}[t]
\centering
\small
\setlength{\tabcolsep}{4pt}
\begin{tabular}{lllllll}
\toprule
Dataset & Comparison & MAPE & $\Delta$ & 95\% CI & $t$-stat & $p$-value \\
\midrule
Airbnb      & $K{=}0 \rightarrow K{=}8$ & $30.31 \rightarrow 27.78$ & -2.53   & [-2.73,-2.33]     & -25.31 & $<.001$ \\
Airbnb      & QR $\rightarrow$ QT       & $27.78 \rightarrow 26.89$ & -0.89   & [-1.01,-0.77]     & -15.03 & $<.001$ \\
StackSample & $K{=}0 \rightarrow K{=}8$ & $266.65 \rightarrow 98.56$ & -168.09 & [-174.56,-161.62] & -50.84 & $<.001$ \\
StackSample & QR $\rightarrow$ QT       & $98.56 \rightarrow 84.30$ & -14.26  & [-17.27,-11.25]   & -9.15  & $<.001$ \\
\bottomrule
\end{tabular}
\caption{Paired statistical significance tests on per-instance avg-MAPE using Qwen3-4B. Comparisons are $K{=}0 \rightarrow K{=}8$ under QR, and QR $\rightarrow$ QT at $K{=}8$. Negative $\Delta$ indicates lower error for the second condition.}
\label{tab:significance-tests}
\end{table*}
We assess whether the improvements from retrieval augmentation and quantile tokens are statistically significant using paired two-sided $t$-tests on per-instance average MAPE. For each instance, we first compute the mean absolute percentage error across the nine evaluation quantiles (10--90), and then compute paired differences between model variants.

Table~\ref{tab:significance-tests} reports results for Qwen3-4B on both datasets. Increasing the number of retrieved neighbors from $K{=}0$ to $K{=}8$ yields statistically significant improvements on both Airbnb and StackSample ($p<0.001$). Similarly, replacing QR with QT at $K{=}8$ also leads to significant gains. The magnitude of improvement is particularly large on StackSample, where retrieval reduces avg-MAPE by 168.09 points, and QT further reduces it by 14.26 points.

Overall, all comparisons are statistically significant with large effect sizes, confirming that the observed improvements are not due to random variation.

\subsection{Autoregressive Baseline}
\label{app:autoregressive}

We evaluate an autoregressive baseline using Claude Sonnet 4 \citep{anthropic2025claude-sonnet4} in a few-shot setting on StackSample. To ensure a controlled comparison, we match the retrieval setup and use $K{=}8$ neighbors. Each prompt includes the query and retrieved examples, and the model is instructed to directly generate the target quantiles.

Due to the brittleness of long-form numeric generation, we restrict the model to output a compact set of 9 quantiles (10--90). As a result, metrics that require dense quantile grids (e.g., RCIW@95 and RCIW@99) are not applicable.

\begin{table}[th]
\centering
\small
\setlength{\tabcolsep}{5pt}
\begin{tabular}{lrrrrrr}
\toprule
\textbf{Model} & \textbf{Method} & \textbf{avg MAPE$\downarrow$} & \textbf{wMAPE$\downarrow$} & \textbf{sMAPE$\downarrow$} & \textbf{CRPSS$\uparrow$} & \textbf{RCIW@90$\downarrow$} \\
\midrule
Claude Sonnet 4 & Autoregressive & 144.87 & 73.36 & 64.71 & 0.3279 & 348.29 \\
Qwen3-4B & QR ($K{=}8$) & 98.56 & 74.97 & 67.19 & 0.3001 & 545.50 \\
Qwen3-4B & QT ($K{=}8$) & \textbf{84.30} & \textbf{73.02} & \textbf{64.86} & \textbf{0.3375} & \textbf{274.20} \\
\bottomrule
\end{tabular}
\caption{Comparison with an autoregressive baseline (Claude Sonnet 4) on the StackSample test split. The autoregressive model outputs 9 quantiles (10--90), so only metrics defined on this subset are reported.}
\label{tab:generative_baseline}
\end{table}

Table~\ref{tab:generative_baseline} compares the autoregressive baseline with fine-tuned models under the same retrieval setting. The autoregressive baseline exhibits substantially higher avg-MAPE, driven by large errors in the upper quantiles. While metrics such as sMAPE, wMAPE, and CRPSS may appear competitive, they primarily reflect central tendency and can mask extreme deviations. In particular, the model tends to underestimate high quantiles, resulting in under-dispersed predictions and poor coverage. Overall, these results indicate that fine-tuned regression models, especially QT, are more reliable for distributional prediction.

\subsection{Cross-Family Validation}
To evaluate whether the proposed method generalizes beyond the Qwen3 model family, we conduct experiments on \texttt{Phi-3-mini-4k-instruct}~\citep{abdin2024phi3technicalreporthighly} and compare QR and QT under the same retrieval setup.

\begin{table*}[thb]
\centering
\small
\setlength{\tabcolsep}{4pt}
\begin{tabular}{lrrrrrrrr}
\toprule
Method & $K$ & avg MAPE$\downarrow$ & wMAPE$\downarrow$ & sMAPE$\downarrow$ & CRPSS$\uparrow$ & RCIW@90$\downarrow$ & RCIW@95$\downarrow$ & RCIW@99$\downarrow$ \\
\midrule
QR & 0 & 255.99 & 75.40 & 68.77 & 0.1795 & 3091.45 & 4847.60 & 5729.65 \\
QR & 8 & 101.63 & 75.68 & 69.64 & 0.3077 & 1600.67 & 2784.72 & 3876.98 \\
QT & 8 & \textbf{82.67} & \textbf{73.36} & \textbf{64.58} & \textbf{0.3469} & \textbf{297.25} & \textbf{333.43} & \textbf{361.66} \\
\bottomrule
\end{tabular}
\caption{Cross-family validation on \texttt{Phi-3-mini-4k-instruct} on StackSample. We compare QR and QT with $K \in \{0,8\}$.}
\label{tab:phi3-cross-family}
\end{table*}

The results in Table~\ref{tab:phi3-cross-family} show consistent trends: retrieval improves the QR baseline, and QT yields further gains in both accuracy and distributional quality. This suggests that the effectiveness of quantile tokens and retrieval augmentation is not specific to a single backbone model.

\end{document}